\documentclass{article}

\usepackage[main, final]{neurips_2026}
\usepackage[ruled,vlined]{algorithm2e}

\usepackage[utf8]{inputenc} 
\usepackage[T1]{fontenc}    
\usepackage{hyperref}       
\usepackage{url}            
\usepackage{booktabs}       
\usepackage{amsfonts}       
\usepackage{nicefrac}       
\usepackage{microtype}      
\usepackage{xcolor}         
\usepackage{amsmath} 
\usepackage{graphicx}
\usepackage{float} 
\usepackage[most]{tcolorbox}
\usepackage{listings}
\usepackage[table]{xcolor}
\tcbset{
  mycodebox/.style={
    enhanced,
    boxrule=0.6pt,
    colback=white,
    colframe=black!60,
    arc=2mm,            
    outer arc=2mm,
    left=2mm,right=2mm,top=1.5mm,bottom=1.5mm,
  }
}

\title{DeCoRL: Decomposed Consistency Reinforcement Learning for Multi-Image Composition}

\author{%
  Zhiqiang Wu\thanks{Work done during internship at Xiaohongshu} \quad Shuang Sun \quad Jiale Zhang \quad Jing Li \quad Yibo Chen \quad Xu Tang \quad Yao Hu \\
  Xiaohongshu Inc. \\
  \texttt{\{wuzhiqiang,sunshuang1,zhangjiale,lijing5\}@xiaohongshu.com} \\
  \texttt{\{zhaohaibo, tangshen, xiahou\}@xiaohongshu.com} 
}

\begin{document}

\maketitle

\begin{figure}[H]
    \vspace{-8mm}
    \centering
    \includegraphics[width=\linewidth,height=\textheight,keepaspectratio]{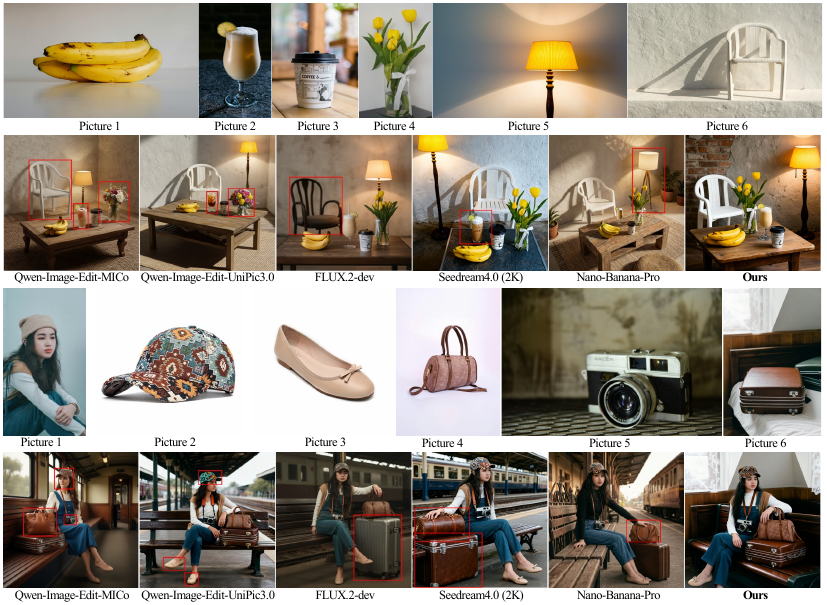}
    \vspace{-4mm}
    \caption{\textbf{
    Qualitative comparisons on two challenging MICo examples, each involving six reference images.}
    Compared with open-source and proprietary baselines, our method better preserves the identities
    and visual attributes of multiple referenced objects in the composed image.
    The {\color{red} red boxes} highlight objects with inconsistent appearances.}
    \label{fig:vis}
\end{figure}

\begin{abstract}
Multi-image composition (MICo) is a critical yet challenging capability for large-scale image editing models, where consistency degrades rapidly as the number of reference images increases. While supervised fine-tuning (SFT) can equip models with basic MICo ability, it provides only implicit supervision on the composed image and often fails to learn reliable object-level correspondence across multiple references. We propose DeCoRL, a reinforcement-learning (RL) framework that improves MICo consistency by decomposing holistic MICo consistency into object-level consistency. DeCoRL first establishes instruction-relevant object correspondences across reference images and localizes the corresponding regions in the generated image. It then uses a VLM-based reward model trained on single-object consistency annotations to produce fine-grained, object-wise feedback, which is aggregated into a multi-object consistency reward to guide RL post-training. We further introduce DeCo-Bench, a benchmark covering compositions from 2 to 6 reference images. Experiments on DeCo-Bench and the public MICo-Bench demonstrate consistent improvements in MICo consistency.

\end{abstract}

\section{Introduction}
Recent instruction-based editing models~\cite{instructpix2pix,longcat,qwen2511,step1x,seedream40,firered,fluxkontext,smartedit,ji2025instruction} have achieved strong performance in instruction following and image quality. However, most open-source systems~\cite{fluxkontext,longcat} are designed for single-image editing and cannot naturally support {multi-image composition} (MICo), where multiple reference images must be jointly composed into a single edited result. Although recent models such as Qwen-Image-Edit-2509/2511~\cite{qwen2511} and FLUX.2-dev~\cite{flux2} begin to support this setting, they still struggle to maintain consistency as the number of reference images increases.

A central challenge in MICo is {cross-image consistency}: the model must integrate multiple visual conditions while preserving the correct identities, attributes, and semantic bindings of multiple entities in the generated image. This is fundamentally harder than single-image editing because the model must resolve many ambiguous many-to-one correspondences: each generated entity may inherit visual properties from a different reference image, and this ambiguity grows quickly with the number of inputs~\cite{zhu2025mde,li2025moedit,zhou2025multi}.

Despite this challenge, most existing MICo approaches primarily rely on large-scale (often synthetic) multi-image editing data and {supervised fine-tuning} (SFT) to endow models with multi-reference capability~\cite{mico,unipics30}. However, SFT typically provides only implicit supervision on the generated image, without explicitly teaching object-level correspondence and binding across references, which often leads to identity confusion or attribute leakage when the number of conditions grows~\cite{zhu2025mde,li2025moedit,zhou2025multi}.

These limitations motivate reinforcement learning (RL) as a post-training paradigm for image editing, since it can directly optimize model preferences via reward signals rather than merely imitate paired targets as in SFT. Recent editing works have similarly adopted RL-style post-training to improve alignment and controllability beyond pure SFT~\cite{editr1,zhao2026imageedit,li2026thinkrl,luo2025editscore,tian2025unigen,guo2026leveraging,wu2026cocoedit,jiang2025t2i}.

The effectiveness of RL hinges on a reliable reward. However, designing such a reward for MICo consistency is particularly challenging: as the number of reference images grows, holistic scoring must jointly resolve ambiguous correspondence and attribute-to-source binding under many competing visual conditions—an ill-posed judgment that remains unreliable even for strong VLM evaluators (e.g., GPT-4o) due to limited cross-image attention~\cite{mico}.

We propose \textbf{DeCoRL} (\textbf{De}composed \textbf{Co}nsistency \textbf{R}einforcement \textbf{L}earning), an RL framework built on a central methodological insight: holistic MICo consistency can be decomposed into tractable consistency checks over instruction-relevant, localizable objects. This decomposition is the core reward-design principle, while the specific evaluator, grounding model, and RL optimizer are implementation choices. We first train a single-object consistency reward model using a vision-language model (VLM)~\cite{radford2021learning,li2022blip,alayrac2022flamingo,liu2023visual}, with chain-of-thought (CoT)~\cite{cot,zhao2026robustness,chen2024measuring,zhang2025improve} supervision, and use its \texttt{yes}/\texttt{no} token probabilities~\cite{chen2025token,liu2026promptecho,chen2026topreward} as a continuous reward signal. We then build an automatic object alignment pipeline: a VLM identifies instruction-relevant objects across the reference images, and Grounding DINO~\cite{dino} localizes them. Based on these aligned objects, we define a \emph{multi-object consistency reward} that {matches generated objects with their corresponding references}, scores each matched pair using \emph{our trained single-object consistency reward model}, and {penalizes missing objects} (i.e., instruction-relevant entities not present in the generated output). This multi-object reward is then used to post-train the editing model via diffusion-based RL.

We further introduce \textbf{DeCo-Bench}, a dedicated benchmark for evaluating MICo consistency. By performing SFT followed by DeCoRL on the Qwen-Image-Edit2511~\cite{qwen2511} model, we observe substantial and consistent improvements in MICo consistency on both DeCo-Bench and the public MICo-Bench, demonstrating the effectiveness of our method.

Our main contributions are summarized as follows:
\begin{itemize}
    \item \textbf{Core insight.} We identify object-level decomposition as a practical reward-design principle for MICo: once instruction-relevant object correspondences are established, consistency can be evaluated locally instead of through a single holistic judgment.

    \item \textbf{DeCoRL framework.} We develop an RL post-training framework that aggregates object-level consistency scores into a multi-object reward, including penalties for missing objects. VLM parsing, Grounding DINO localization, and CoT-trained reward modeling provide one modular instantiation of this framework.

    \item \textbf{Benchmark and evaluation.} We introduce DeCo-Bench (2--6 reference images) and demonstrate consistent improvements on both DeCo-Bench and the independently developed MICo-Bench.
\end{itemize}

\section{Related Work}

\paragraph{Single-Image Editing.}
Instruction-based image editing has advanced rapidly with the development of text-to-image (T2I) foundation models~\cite{sd3,longcat,qwen2511,flux1,hunyuanimage,zimage,seedream20,seedream30}. Representative open-source models (LongCat-Image-Edit~\cite{longcat} and FLUX.1-Kontext~\cite{fluxkontext}) achieve strong performance in instruction following and image quality. However, these methods are primarily designed for single-image editing and do not support MICo.

\paragraph{Multi-Image Composition.}
Multi-image composition (MICo) requires jointly integrating several reference images into a single generated result, making it substantially more challenging than single-image editing. Recent open-source models such as Qwen-Image-Edit~\cite{qwen2511}, FLUX.2-dev~\cite{flux2}, and FireRed-Image-Edit~\cite{firered} have begun to support this setting, while advanced closed-source systems such as Nano-Banana-Pro~\cite{nanopro} and Seedream 4.0~\cite{seedream40} further highlight its practical importance. However, maintaining MICo consistency remains difficult, especially as the number of reference images increases.

Most recent MICo progress relies on large-scale (often synthetic) data construction followed by supervised fine-tuning (SFT), which can efficiently imitate demonstrations but typically supervises only the final composed image and provides limited explicit signals for learning cross-image correspondence and consistency relations. As a result, MICo consistency often degrades in harder cases with many reference images. Notably, this issue also appears in strong closed-source systems (e.g., Nano-Banana-Pro~\cite{nanopro} and Seedream 4.0~\cite{seedream40}), which sometimes produce inconsistent compositions, e.g., object identities or attributes may change in complex multi-reference settings.

\paragraph{Reinforcement-Learning for Image Editing.}
Reinforcement learning (RL) has become an effective post-training tool for diffusion-based image editing, and has been increasingly adopted in recent open-source systems. For example, Skywork UniPic 2.0 explicitly incorporates online RL to improve unified multimodal generation and editing capability~\cite{unipics20}, while LongCat-Image-Edit~\cite{longcat}, Qwen-Image-Edit~\cite{qwen2511}, Edit-R1~\cite{editr1}, and FireRed-Image-Edit~\cite{firered} also leverage preference optimization or reward-driven training (e.g., DPO~\cite{dpo}, GRPO~\cite{grpo}, Flow-GRPO~\cite{flowgrpo}, and Diffusion NFT~\cite{nft}) to enhance image quality, instruction adherence, and controllability. 

These successful practices suggest that RL provides a promising direction for improving editing models beyond standard SFT objectives. Motivated by this, we propose DeCoRL, which leads to substantial improvements in MICo consistency.

\section{Method}
\label{sec:method}

\begin{figure}
    \centering
    \includegraphics[width=0.98\linewidth]{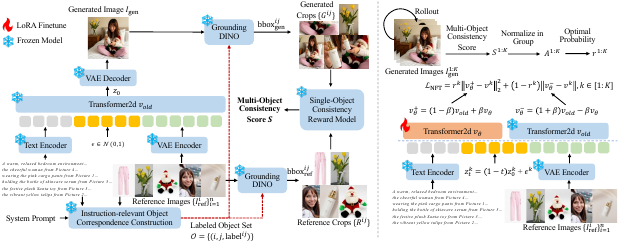}
    \caption{\textbf{Overview of DeCoRL.} DeCoRL improves MICo consistency via decomposed, object-level RL. We use a VLM to parse the instruction and identify instruction-relevant objects in each reference image, then localize them with Grounding DINO to obtain object crops. For a generated image, we detect the corresponding objects, score each reference--generated object pair with a trained single-object consistency reward model, aggregate the scores into a multi-object consistency reward, and finally optimize the transformer2d model with Diffusion NFT~\cite{nft}.}
    \label{fig:framework}
\end{figure}

We present \textbf{DeCoRL} (see Figure~\ref{fig:framework}), an RL-based post-training framework for improving consistency in MICo. The key idea of DeCoRL is to avoid directly scoring MICo consistency between multiple reference images and a generated image, which is highly ambiguous and difficult to evaluate reliably (as stated in~\cite{mico}), and instead introduce a principled decomposition into object-level consistency evaluation. This design is motivated by the observation that, once object correspondence is established, single-object consistency is more tractable to evaluate than global MICo consistency. The decomposed reward primarily targets the identity and visual attributes of localizable, instruction-relevant objects; the particular grounding and scoring components are modular.

\subsection{Task Definition}
\label{subsec:task_overview}

In MICo, the input consists of a textual editing instruction $c$ and a set of reference images $\mathcal{I}_{\mathrm{ref}} = \{I_{\mathrm{ref}}^i\}_{i=1}^{n}$, 
and the goal is to generate an edited image $I^{\mathrm{gen}}$ that follows the instruction while preserving relevant consistency from the reference images.

\subsection{Multi-Image Supervised Fine-Tuning}
\label{subsec:sft_init}

We adopt Qwen-Image-Edit2511~\cite{qwen2511} as the base image editing model. Since the original model does not reliably support generation with more than three reference images, we first perform supervised fine-tuning (SFT) covering 1--6 reference images. For multi-image supervision, we combine MICo-150K~\cite{mico} and UniPic 3.0~\cite{unipics30}, yielding approximately 470K training examples across this reference range.

As the number of reference images increases, the number of visual tokens grows rapidly. To keep the total visual input within the model capacity, we resize all reference images under a shared total-pixel budget while preserving their aspect ratios. Let the $i$-th reference image $I_{\mathrm{ref}}^i$ have width $w^i$ and height $h^i$. We first compute the total reference area
$
U = \sum_{i=1}^{n} w^i h^i,
$
and define a shared scaling factor
$
f = \sqrt{\frac{P}{U}},
$
where $P$ is the target total area budget (e.g., $1024^2$). Each reference image is then resized as
\begin{equation}
\hat{I}_{\mathrm{ref}}^i
=
\mathrm{Resize}\!\left(
I_{\mathrm{ref}}^i,\;
\mathrm{round}(w^i f),\;
\mathrm{round}(h^i f)
\right),
\label{eq:resize_ref}
\end{equation}
so that aspect ratios and relative scales are preserved.

\paragraph{Latent Encoding.}
We encode the resized references and the target image $I^{\mathrm{tgt}}$ into latent features using a VAE encoder $\mathcal{E}$~\cite{vae}:
\begin{equation}
z_{\mathrm{ref}}^i = \mathcal{E}\!\left(\hat{I}_{\mathrm{ref}}^i\right),
\qquad
z_{\mathrm{tgt}} = \mathcal{E}\!\left(I_{\mathrm{tgt}}\right).
\end{equation}
We then concatenate each reference latent feature into a whole latent feature:
\begin{equation}
z_\mathrm{ref} = [z_{\mathrm{ref}}^1;z_{\mathrm{ref}}^2;\cdots;z_{\mathrm{ref}}^n]
\end{equation}

\paragraph{Training Objective.}
We train the transformer2d model with a flow-matching~\cite{fm} objective. We first sample a scalar time step $t \in (0,1)$ from a logit-normal distribution~\cite{sd3}:
$
t = \sigma(u), u \sim \mathcal{N}(0,1)
$
where $\sigma(\cdot)$ is the sigmoid function. We then sample Gaussian noise
$
\epsilon \sim \mathcal{N}(0, I),
$
with the same shape as $z_{\mathrm{tgt}}$, and construct the noisy latent feature at timestep $t$:
\begin{equation}
z_t = (1-t)\,z_{\mathrm{tgt}} + t \epsilon.
\label{eq:noised_target}
\end{equation}
The transformer2d model~\cite{sd3} predicts the velocity field for the target latent feature:
\begin{equation}
v_\theta = \operatorname{Transformer2d}(z_t, z_{\mathrm{ref}}, t, c).
\end{equation}
Under the linear interpolation in Eq.~\eqref{eq:noised_target}, the ground-truth velocity is
$
v_t = \frac{dz_t}{dt} = \epsilon - z_{\mathrm{tgt}}.
$
We optimize the flow-matching loss
\begin{equation}
\mathcal{L}_{\mathrm{FM}} = \left\| v_\theta - v_t \right\|_2^2.
\end{equation}

\paragraph{Role of SFT.}
This stage provides a strong initialization for multi-reference editing, but it does not explicitly supervise object correspondences across reference images. Consequently, the editor may still omit requested objects, confuse their identities, or fail to preserve local attributes, motivating the object-level consistency rewards introduced next.

\subsection{Single-Object Consistency Reward model}
\label{subsec:single_obj_reward}

We train a single-object consistency reward model on PaCO-RL-Dataset~\cite{pacorl} under a chain-of-thought (CoT) evaluation protocol, since \texttt{yes}/\texttt{no}-only supervision is low-entropy and prone to shortcut overfitting~\cite{cot,zhao2026robustness,chen2024measuring,zhang2025improve}. We adopt Qwen3-VL-32B~\cite{qwen3vl} as the reward backbone and fine-tune it to answer binary consistency questions.

\paragraph{Training.}
Each training example contains a reference crop $\mathbf{x}_{\mathrm{ref}}$, a generated crop $\mathbf{x}_{\mathrm{gen}}$, and a textual query $q$ about their identity and visual-attribute consistency. We fine-tune the VLM with standard autoregressive cross-entropy on annotated CoT responses containing an explanation and a binary decision (\texttt{yes}/\texttt{no}).

\paragraph{Inference.}
We decode up to $T_{\mathrm{check}}$ steps and identify the first step $t^\star$ at which the highest-logit vocabulary token is \texttt{yes} or \texttt{no}. Denoting their logits by $\ell^{(t^\star)}_{\texttt{yes}}$ and $\ell^{(t^\star)}_{\texttt{no}}$, we compute the single-object consistency score as
\begin{equation}
s_{\mathrm{obj}} =
\begin{cases}
\sigma\!\left(\ell^{(t^\star)}_{\texttt{yes}}-\ell^{(t^\star)}_{\texttt{no}}\right), & \text{if }t^\star\text{ exists},\\[4pt]
0, & \text{otherwise},
\end{cases}
\label{eq:single_object_score}
\end{equation}
where $\sigma(\cdot)$ is the sigmoid function. The resulting score in $[0,1]$ provides continuous object-level feedback.

\subsection{Instruction-Relevant Object Correspondences}
\label{subsec:vlm_grounding}
\begin{figure}
    \centering
    \includegraphics[width=1\linewidth]{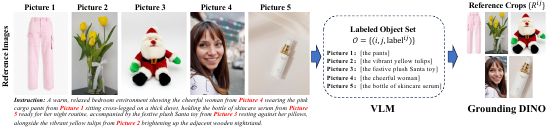}
    \caption{\textbf{Instruction-relevant Object Correspondences.} A VLM jointly interprets the instruction
and multiple reference images, identifies instruction-relevant objects, and produces a labeled object set
for Grounding DINO to obtain the reference crops.}
    \label{fig:vlm_2}
\end{figure}
To make object-level rewards applicable to MICo, we first construct
{instruction-relevant object correspondences} from the instruction and reference images, as illustrated in Figure~\ref{fig:vlm_2}. This step converts holistic MICo into a set of aligned object instances, enabling reliable evaluation with our trained single-object consistency reward model.

\paragraph{Instruction-aware Object Set.}
Given an instruction $c$ and reference images
$\mathcal{I}_{\mathrm{ref}}=\{I^i_{\mathrm{ref}}\}_{i=1}^{n}$, we use a VLM to identify the objects explicitly required by $c$ and to assign an open-vocabulary label to each object instance in each reference
image. We denote the resulting labeled object set as
\[
\mathcal{O} = \{(i,j,\mathrm{label}^{ij})\},
\]
where $(i,j)$ indexes the $j$-th instruction-relevant object in the
$i$-th reference image. Focusing on $\mathcal{O}$ prevents background
clutter or instruction-irrelevant entities from affecting reward
computation.

\paragraph{Open-vocabulary Grounding in References.}
For each $(i,j,\mathrm{label}^{ij})\in \mathcal{O}$, we apply Grounding DINO on $I^i_{\mathrm{ref}}$ with the text prompt $\mathrm{label}^{ij}$ and obtain a set of candidate detections with confidence scores. We keep only the highest-confidence (top-1) detection (after standard post-processing such as thresholding and NMS), and denote its bounding box as
$\mathrm{bbox}^{ij}_{\mathrm{ref}}$. The corresponding reference crop is
\begin{equation}
R^{ij}=\mathrm{Crop}(I^i_{\mathrm{ref}}, \mathrm{bbox}^{ij}_\mathrm{ref}).
\end{equation}
These grounded crops serve as the spatial anchors for subsequent
object-level consistency scoring.

\subsection{Multi-Object Consistency Reward}
\label{subsec:multi_obj_reward}

\begin{figure}
    \centering
    \includegraphics[width=1\linewidth]{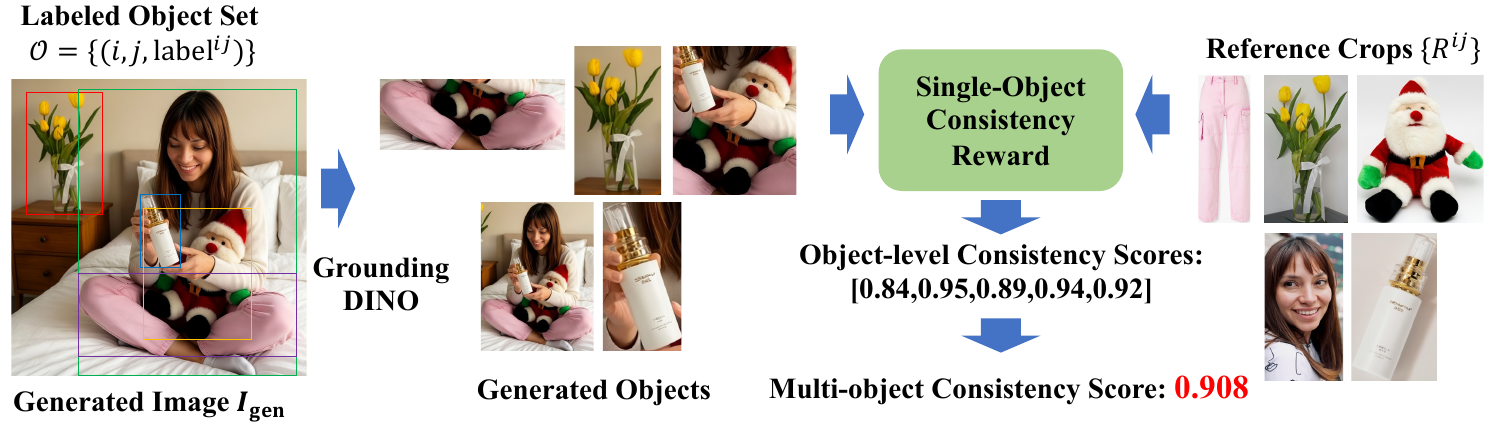}
    \caption{\textbf{Multi-object Consistency Reward.} Object-level consistency scores are computed for aligned reference-generated object pairs and then aggregated into a single-object consistency reward.}
    \label{fig:multi_reward}
\end{figure}

We now construct a multi-object consistency reward for MICo by composing the single-object consistency scores, as shown in ~\ref{fig:multi_reward}.

\paragraph{Generated-object Grounding and Matching.}
For each object label $\mathrm{label}^{ij}$ identified in the reference images,
we also apply Grounding DINO on $I_{\mathrm{gen}}$ with the text prompt
$\mathrm{label}^{ij}$ and keep the highest-confidence (top-1) detected bounding
box $\mathrm{bbox}^{ij}_{\mathrm{gen}}$. The corresponding generated crop is
\begin{equation}
G^{ij} = \mathrm{Crop}(I_{\mathrm{gen}}, \mathrm{bbox}^{ij}_\mathrm{gen}).
\end{equation}

\paragraph{Per-Object Reward.}
For each pair $(R^{ij}, G^{ij})$, we resize both crops while preserving aspect ratio and feed them into the single-object consistency reward model. If an object exists in the reference image but is not detected in the generated image, we assign a reward of zero to explicitly penalize missing objects:
\begin{equation}
s^{ij} =
\begin{cases}
s_{\mathrm{obj}}\!\left(\mathrm{Resize}(R^{ij}), \mathrm{Resize}(G^{ij})\right), & G^{ij} \neq \emptyset,\\[4pt]
0, & \text{otherwise}.
\end{cases}
\label{eq:per_object_reward}
\end{equation}

\paragraph{Reward Aggregation.}
Let $N$ denote the number of valid reference objects in the sample. We aggregate the object-level consistency scores into a multi-object consistency score:
\begin{equation}
S = \dfrac{1}{N}\sum_{i,j} s^{ij},
\label{eq:multi_object_reward}
\end{equation}
where $N>0$ denotes the number of valid instruction-relevant reference objects. The decomposition makes each required object's consistency explicitly measurable and assigns zero reward to missing detections, rather than relying on a single undifferentiated whole-image judgment. The mean is deliberately simple; it does not by itself guarantee balanced improvement across objects.

\subsection{Diffusion NFT Post-Training}
\label{subsec:diffusion_nft}


Using the multi-object reward defined above, we further optimize the editing model through Diffusion NFT~\cite{nft}, a practical choice for diffusion RL post-training. The decomposition-based reward is not specific to this optimizer. The objective is to increase the likelihood of reward-improving predictions while keeping the updated policy close to a reference model for stability.

\paragraph{Reward Normalization and Policy Optimization.}
Following Diffusion NFT~\cite{nft}, for $K$ samples generated under the same condition $c$, let $S^k$ denote the multi-object reward of sample $k$, and let $\mu$ and $\sigma$ be the group mean and standard deviation. We normalize and clip the relative advantages, mapping them to weights in $[0,1]$:
\begin{equation}
r^k = \frac{1}{2} + \frac{1}{2A_{\max}}\,
\mathrm{clip}\!\left(\frac{S^k-\mu}{\sigma+\epsilon},-A_{\max},A_{\max}\right),
\label{eq:ri}
\end{equation}
where $\epsilon>0$ ensures numerical stability and $A_{\max}>0$ sets the clipping bound. These weights balance the positive and negative optimization directions:
\begin{equation}
\mathcal{L}_{\mathrm{NFT}} = \mathbb{E}_{k,t}\!\left[
r^k \|v_{\theta}^{+}-v^k\|_2^2
+(1-r^k)\|v_{\theta}^{-}-v^k\|_2^2
\right],
\label{eq:single_policy_loss}
\end{equation}
where $v^k=\epsilon^k-z_{\mathrm{gen}}^k$ with $z_{\mathrm{gen}}^k=\mathcal{E}(I_{\mathrm{gen}}^k)$, and $v_\theta^{+}=(1-\beta)v_{\mathrm{old}}+\beta v_\theta$ and $v_\theta^{-}=(1+\beta)v_{\mathrm{old}}-\beta v_\theta$ denote the positive and negative velocities, respectively. Here $\beta$ is a hyperparameter, and the loss is evaluated over sampling timesteps $t$ for samples generated by the old policy.

To prevent the policy from drifting too far from the initial model, we use reference regularization~\cite{kl1,kl2} that penalizes differences between the current velocity prediction $v_\theta$ and the reference prediction $v_{\mathrm{ref}}$. The final DeCoRL objective is
\begin{equation}
\mathcal{L} = \lambda_{\mathrm{NFT}}\mathcal{L}_{\mathrm{NFT}}
+\lambda_{\mathrm{Reg}}\,
\underbrace{\mathbb{E}_{k,t}\!\left[\|v_\theta-v_{\mathrm{ref}}\|_2^2\right]}_{\mathcal{L}_{\mathrm{Reg}}},
\label{eq:final_loss}
\end{equation}
where $\lambda_{\mathrm{NFT}}$ and $\lambda_{\mathrm{Reg}}$ are hyperparameters. In practice, to mitigate reward hacking, we additionally incorporate a general-purpose VLM (Qwen3.5-122B-10A)~\cite{qwen35} reward that penalizes unnatural outputs, structural distortions, and compositing artifacts. This complementary signal also discourages globally implausible object--subject interactions that might be missed by local crop-based consistency scoring.

\section{Experiments}
\label{sec:experiments}
We provide further implementation details, including our trained single-object consistency reward model, in Appendix~\ref{app:details}. Additional diagnostic evaluations and a discussion of scope are provided in Appendix~\ref{app:additional_validation}.

\subsection{Training Setup.}
\textbf{SFT.} We fine-tune Qwen-Image-Edit2511 on the multi-image datasets MICo-150K~\cite{mico} and UniPic 3.0~\cite{unipics30} described in Section~\ref{subsec:sft_init}, together with Nano-consistent-150K~\cite{nano150k} for single-image editing. We train for 24{,}000 steps on 32 H800 GPUs with a total batch size of 128, using AdamW ($\beta_1=0.9$, $\beta_2=0.999$, $\epsilon=10^{-8}$, weight decay $10^{-2}$) and a learning rate of $2\times10^{-5}$.

\textbf{RL.} We initialize from the SFT checkpoint and perform Diffusion NFT post-training on MICo-150K~\cite{mico} using LoRA~\cite{lora} ($r=32$). Training uses 32 H800 GPUs, a group size of 16, and AdamW ($\beta_1=0.9$, $\beta_2=0.999$, $\epsilon=10^{-10}$, weight decay $10^{-4}$) with a learning rate of $3\times10^{-4}$.

\subsection{Benchmarks}
\paragraph{DeCo-Bench.}
To evaluate consistency in MICo, we construct \textbf{DeCo-Bench}, a benchmark covering MICo tasks with 2--6 reference images. Each setting contains 40 test cases (a total of 200 cases), and all reference images are real-world photographs. We provide details of DeCo-Bench in Appendix~\ref{app:deco_bench}.

\paragraph{MICo-Bench.}
To further assess overall performance, we also evaluate our method on the public {MICo-Bench}. Following its standard protocol, we additionally report category-level results under {Object Centric} (138 cases), {Human Centric} (168 cases), {HOI} (291 cases), and {De\&Re} (300 cases). We additionally report results on {Semantic Consistency (SC)}, including Prompt Following (PF) and Subject Resemblance (SR), as well as {Perceptual Quality (PQ)}, including Naturalness and Artifact. We provide details of MICo-Bench in Appendix~\ref{app:mico_bench}.

\begin{table}
    \centering
    \caption{Results and ablations on DeCo-Bench.}
    \resizebox{\linewidth}{!}{%
    \begin{tabular}{lcccccc}
        \toprule
        \textbf{Model} & \textbf{2 images$\uparrow$} & \textbf{3 images$\uparrow$} & \textbf{4 images$\uparrow$} & \textbf{5 images$\uparrow$} & \textbf{6 images$\uparrow$} & \cellcolor{gray!20}\textbf{Overall$\uparrow$}\\
        \midrule
        Qwen-Image-Edit-MICo~\cite{mico} & 6.90 & 5.27 & 4.85 & 4.79 & 4.06 & \cellcolor{gray!20}5.17 \\
        Qwen-Image-Edit-UniPic3.0~\cite{unipics30} & 7.82 & 6.23 & 5.73 & 5.24 & 5.11 & \cellcolor{gray!20}6.02 \\
        FLUX.2-dev~\cite{flux2} & 8.10 & 7.26 & 7.10 & 6.76 & 6.28 & \cellcolor{gray!20}7.10 \\
        \midrule
        Seedream4.0 (2K)~\cite{seedream40} & 8.17 & 8.17 & 7.50 & 7.49 & 7.47 & \cellcolor{gray!20}7.76 \\
        Nano-Banana-Pro~\cite{nanopro} & \textbf{9.26} & \textbf{8.67} & \textbf{8.77} & \textbf{8.58} & \textbf{8.18} & \cellcolor{gray!20}\textbf{8.69} \\
        \midrule
        Qwen-Image-Edit2511 (baseline)~\cite{qwen2511} & 8.41 & 6.99 & 7.09 & 4.98 & 3.64 & \cellcolor{gray!20}6.22 \\
        + SFT & 8.34 & 7.35 & 6.80 & 6.60 & 6.15 & \cellcolor{gray!20}7.05 \\
        + DeCoRL (ours) & \underline{8.91} & \underline{8.36} & \underline{8.05} & \underline{7.88} & \underline{7.61} & \cellcolor{gray!20}\underline{8.16} \\
        \bottomrule
    \end{tabular}
    }
    \label{tab:tb1}
\end{table}

\begin{table}
    \centering
    \caption{Results and ablations on MICo-Bench.}
    \resizebox{\linewidth}{!}{%
    \begin{tabular}{lccccc}
        \toprule
         \textbf{Model} & \textbf{Object Centric$\uparrow$} & \textbf{Human Centric$\uparrow$} & \textbf{HOI$\uparrow$} & \textbf{De\&Re$\uparrow$} & \cellcolor{gray!20}\textbf{Overall$\uparrow$}\\
         \midrule
         Qwen-Image-Edit-MICo~\cite{mico} & 52.38 & 21.11 & 34.95 & 37.42 & \cellcolor{gray!20}35.86 \\
         Qwen-Image-Edit-UniPic3.0~\cite{unipics30} & \textbf{67.94} & 44.79 & 28.24 & \underline{59.28} & \cellcolor{gray!20}47.90 \\
         FLUX.2-dev~\cite{flux2} & 41.21 & 45.10 & \underline{49.86} & 55.71 & \cellcolor{gray!20}49.55 \\
         \midrule
         Nano-Banana~\cite{nano}  & 48.01 & 41.79 & 49.64 & 49.44 & \cellcolor{gray!20}47.83 \\
         Nano-Banana-Pro~\cite{nanopro}  & 50.59 & \underline{54.75} & \textbf{50.21} & 52.13 & \cellcolor{gray!20}\underline{51.76} \\
         \midrule
         Qwen-Image-Edit2511 (baseline)~\cite{qwen2511} & 27.41 & 28.89 & 19.59 & 20.60 & \cellcolor{gray!20}22.90 \\
         + SFT & 39.60 & 30.49 & 32.41 & 45.50 & \cellcolor{gray!20}37.58 \\
         + DeCoRL (ours) & \underline{53.47} & \textbf{56.21} & 42.49 & \textbf{64.42} & \cellcolor{gray!20}\textbf{54.03} \\
         \bottomrule
    \end{tabular}%
    }
    \label{tab:tb2}
\end{table}

\begin{table}
    \centering
    \caption{Ablations on four MICo-Bench metrics.}
     \resizebox{\linewidth}{!}{%
    \begin{tabular}{lcccc}
        \toprule
         \textbf{Model} & \textbf{Prompt Following (PF)$\uparrow$}  & \textbf{Subject Resemblance (SR)$\uparrow$} & \textbf{Naturalness$\uparrow$} & \textbf{Artifacts$\uparrow$} \\
         \midrule
         Qwen-Image-Edit2511 (baseline) & 6.1495 & 5.2378 & 5.0817 & 4.1208 \\
         + SFT & \textbf{8.7586} & \underline{7.0080} & \underline{6.5355} & \underline{5.7574} \\
         + DeCoRL & \underline{8.6865} & \textbf{7.6393} & \textbf{8.0123} & \textbf{6.6362} \\
        \bottomrule
    \end{tabular}%
    }
    \label{tab:tb4}
\end{table}

\begin{table}
    \centering
    \caption{Ablations of vanilla vs. trained VLM rewards on DeCo-Bench.}
    \resizebox{\linewidth}{!}{%
    \begin{tabular}{lcccccc}
        \toprule
        \textbf{Model} & \textbf{2 images$\uparrow$} & \textbf{3 images$\uparrow$} & \textbf{4 images$\uparrow$} & \textbf{5 images$\uparrow$} & \textbf{6 images$\uparrow$} & \cellcolor{gray!20}\textbf{Overall$\uparrow$}\\
        \midrule
        Qwen-Image-Edit2511 (baseline)~\cite{qwen2511} + SFT & 8.34 & 7.35 & 6.80 & 6.60 & 6.15 & \cellcolor{gray!20}7.05 \\
        \midrule
        w/ \textbf{vanilla} VLM reward & 8.54 & 7.49 & 6.42 & 6.90 & 6.78 & \cellcolor{gray!20}7.21 \\
        w/ \textbf{trained} VLM reward (ours) & \textbf{8.91} & \textbf{8.36} & \textbf{8.05} & \textbf{7.88} & \textbf{7.61} & \cellcolor{gray!20}\textbf{8.16} \\
        \bottomrule
    \end{tabular}
    }
    \label{tab:tb3}
\end{table}

\begin{figure}
    \centering
    \includegraphics[width=0.98\linewidth]{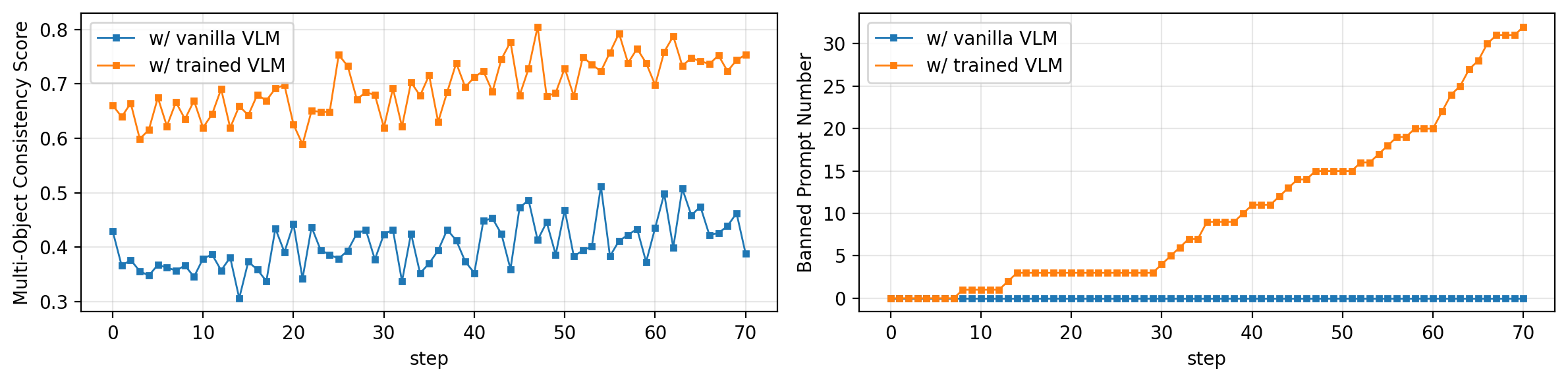}
    \caption{Training dynamics of multi-object consistency score and banned prompt count (vanilla vs. trained VLM).}
    \label{fig:reward_compare}
\end{figure}

\subsection{Comparative Results}

\paragraph{Results on DeCo-Bench.} As shown in Table~\ref{tab:tb1}, performance generally drops as the number of reference images increases (2$\rightarrow$6), consistent with the increased difficulty of MICo consistency. Our method achieves an {overall} score of \textbf{8.16}, {clearly outperforming all open-source baselines}, including Qwen-Image-Edit-MICo (5.17), Qwen-Image-Edit-UniPic (6.02), and FLUX.2-dev (7.10), and also surpassing the proprietary Seedream4.0 (2K) ({7.76}). While Nano-Banana-Pro remains the strongest model overall (8.69), our results indicate that we are {closing the gap toward it}. Despite the remaining difference, the consistent improvements over all other models validate the effectiveness of our method.

\paragraph{Results on MICo-Bench.} Table~\ref{tab:tb2} reports results on the public MICo-Bench. Our method achieves a state-of-the-art overall score of \textbf{54.03}, outperforming both open-source baselines (Qwen-Image-Edit-MICo/UniPic3.0 and FLUX.2-dev) and strong proprietary systems (Nano-Banana and Nano-Banana-Pro).

\subsection{Ablation Study}
\paragraph{Ablation Study on DeCo-Bench.}
Table~\ref{tab:tb1} shows that the {baseline} model struggles severely when the number of reference images increases, with a sharp drop at 5--6 images (4.98 and 3.64). In our tests, this often corresponds to qualitative “collapse” cases where the model fails to preserve key attributes consistently. After {SFT}, consistency improves across all settings (overall: 6.22$\rightarrow$7.05), indicating that supervised training helps the model acquire basic MICo capability.

Applying {DeCoRL} further yields {consistent and substantial gains} from 2 to 6 images (overall: 7.05$\rightarrow$\textbf{8.16}), with especially notable improvements under harder conditions (e.g., 6 images: 6.15$\rightarrow$\textbf{7.61}). These results suggest that while SFT can teach the model to perform MICo, it remains difficult for the model to robustly learn MICo consistency; our DeCoRL is effective in strengthening this consistency across varying numbers of inputs.

\paragraph{Ablation Study on MICo-Bench.}
As shown in Table~\ref{tab:tb2}, our method brings substantial improvements over both the {baseline} and {baseline+SFT} on MICo-Bench. Specifically, DeCoRL boosts the {overall} score from 22.90 (baseline) and 37.58 (+SFT) to \textbf{54.03}, with the most significant gain on Human Centric (28.89$\rightarrow$30.49$\rightarrow$\textbf{56.21}) and {De\&Re} (20.60$\rightarrow$45.50$\rightarrow$\textbf{64.42}).

We further analyze MICo-Bench using four metrics in Table~\ref{tab:tb4}. Compared with the SFT model, DeCoRL yields clear gains in {Subject Resemblance (SR)} (7.01$\rightarrow$\textbf{7.64}), {Naturalness} (6.54$\rightarrow$\textbf{8.01}), and {Artifacts} (5.76$\rightarrow$\textbf{6.64}), indicating better identity preservation and visual quality with fewer artifacts. Meanwhile, {Prompt Following (PF)} is slightly lower than the SFT variant (8.76$\rightarrow$8.69), suggesting that the improvements are not merely due to reward hacking that over-optimizes a single criterion. Notably, SFT consistently improves all four metrics over the baseline, largely because it strengthens basic multi-image editing ability, reducing failure cases and improving instruction following in multi-image settings.

\paragraph{Ablation Study of Single-Object Consistency Reward Model on DeCo-Bench.} 
We provide an ablation study of our single-object consistency reward model mentioned in Section~\ref{subsec:single_obj_reward}, as presented in Table~\ref{tab:tb3}.
Using a vanilla VLM yields only marginal gains over SFT (overall 7.05$\rightarrow$7.21), suggesting limited ability to provide reliable supervision for MICo consistency. In contrast, our CoT-trained VLM substantially improves performance across 2--6 images, reaching \textbf{8.16} overall.

Figure~\ref{fig:reward_compare} further supports this finding. With the vanilla VLM reward, the multi-object consistency score remains mostly below 0.5 with limited improvement, indicating a weak optimization signal. In contrast, our CoT-trained VLM reward stays above 0.6 and steadily increases during training.
The banned prompt count is also informative: a prompt is banned when its samples already obtain high and similar rewards, meaning the model has learned this reward signal. It remains zero for the vanilla VLM, while it increases over time for the CoT-trained VLM, showing that more prompts receive stable and well-learned reward signals. Overall, these results show that the CoT-trained VLM provides a more reliable and informative reward than a vanilla VLM, enabling effective DeCoRL optimization and improving MICo consistency.

\subsection{Visualization}
Figure~\ref{fig:vis} presents two examples of MICo, each consisting of six images. It can be observed that both the open-source model and the closed-source Seedream 4.0 exhibit relatively poor consistency. Even the strongest closed-source model, Nano-Banana-Pro, shows objects with noticeable inconsistency under multi-reference settings. In contrast, our model maintains better consistency in both cases. We also provide more visualizations of our method on DeCo-Bench and MICo-Bench in Appendix~\ref{sec:app_vis}.

\section{Conclusion}
\subsection{Summary}

In this work, we introduce DeCoRL, a reinforcement-learning (RL) post-training framework that improves multi-image composition (MICo) consistency through object-level reward decomposition. DeCoRL establishes instruction-relevant object correspondences, localizes matched regions, and aggregates fine-grained single-object consistency scores into a multi-object reward for optimization. We further propose DeCo-Bench, covering MICo tasks with 2--6 reference images. Experiments on DeCo-Bench and MICo-Bench demonstrate consistent improvements, particularly in challenging many-reference settings, showing the effectiveness of decomposed consistency rewards for scalable MICo post-training.

\subsection{Discussion and Future Directions}
\label{subsec:method_discussion}

\paragraph{From many references to object-wise evaluation.}
DeCoRL is built on the principle that a difficult holistic MICo consistency judgment can be decomposed into more tractable checks. Instead of evaluating all references against the generated image jointly, we first establish instruction-relevant object correspondences, then score each matched reference--generated object pair individually, and finally aggregate these scores into a multi-object reward. This turns a many-reference judgment into a collection of simpler object-level evaluations while retaining a joint objective of preserving multiple subjects. The central methodological contribution is this decomposition-based reward design, rather than a particular VLM, grounding model, or RL optimizer.

\paragraph{Generality and future directions.}
Our CoT-trained VLM is one instantiation of the single-object evaluator. In principle, any reliable evaluator that produces comparable object-level consistency scores can be integrated into the same framework; the reward design is also conceptually independent of Diffusion NFT. This modularity suggests opportunities for other evaluators and editing backbones, although such extensions require empirical validation. The current method targets localizable objects and depends on accurate grounding and matching. Future directions include improving correspondence and robustness to imperfect crops, exploring adaptive aggregation to better protect low-scoring objects, and extending the reward to relational, interaction-level, and other non-localizable consistency. Scaling beyond six references additionally requires suitable training data and further evaluation.

\small
\bibliographystyle{plain}
\bibliography{reference}

\appendix
\newpage
\section{Appendix}

\subsection{Training Details}
\label{app:details}

\paragraph{Single-object consistency reward model.}
We train Qwen3-VL-32B-Instruct~\cite{qwen3vl} on PaCO-RL-Dataset~\cite{pacorl} with LoRA rank 32 and a learning rate of $1\times10^{-4}$ for 2{,}200 steps on 8 H800 GPUs.

\paragraph{Supervised fine-tuning.}
We fine-tune Qwen-Image-Edit-2511, our baseline model, on Nano-consistent-150k~\cite{nano150k}, MICo-150K~\cite{mico}, and UniPic 3.0~\cite{unipics30}. We use AdamW with a learning rate of $2\times10^{-5}$, $\beta_1=0.9$, $\beta_2=0.999$, $\epsilon=1\times10^{-8}$, and weight decay $1\times10^{-2}$. Training uses 32 H800 GPUs, a total batch size of 128, and runs for 24{,}000 steps.

\paragraph{RL post-training.}
Starting from the SFT checkpoint, we further train on MICo-150K with LoRA~\cite{lora} rank $r=32$. RL uses 32 H800 GPUs, group size 16, 8 prompts per sample, and AdamW with learning rate $3\times10^{-4}$, $\beta_1=0.9$, $\beta_2=0.999$, $\epsilon=1\times10^{-10}$, and weight decay $1\times10^{-4}$. To reduce reward hacking, we add a general-purpose VLM reward from Qwen3.5-122B-10A~\cite{qwen35} to penalize unnatural outputs, structural distortions, and compositing artifacts. We set $\lambda_\mathrm{NFT}=1$, $\lambda_\mathrm{Reg}=10^{-4}$, and $\beta=1$. See Algorithm~\ref{alg:decorl_compact} for pseudocode.

\begin{algorithm}[h]
\caption{DeCoRL Diffusion NFT Post-Training}
\label{alg:decorl_compact}
\KwIn{Dataset $\mathcal{D}$, current policy $v_\theta$, old policy $v_{\mathrm{old}}$, reference policy $v_{\mathrm{ref}}$}
\While{not converged}{
    Sample $(c,\mathcal{I}_{\mathrm{ref}}) \sim \mathcal{D}$\;
    Generate a group of samples $\{I^k_{\mathrm{gen}}\}_{k=1}^K$ using $v_{\mathrm{old}}$\;
    Compute rewards $\{S_k\}_{k=1}^K$ with the multi-object consistency reward in Eq.~\eqref{eq:multi_object_reward} and the general VLM reward\;
    Normalize rewards into weights $\{r_k\}_{k=1}^K$ using Eq.~\eqref{eq:ri}\;
    \For{each sample $k$}{
        \For{each sample timestep $t$}{
            Sample $(z_t^k,t)$ using Eq.~\eqref{eq:noised_target}\;
            Compute $v_\theta$, $v_{\mathrm{old}}$, and $v_{\mathrm{ref}}$\;
            Construct $v^+_\theta$ and $v^-_\theta$ using Eqs.~\eqref{eq:single_policy_loss}\;
            Compute $\mathcal{L}_\mathrm{NFT}$ using Eq.~\eqref{eq:single_policy_loss}\;
            Compute $\mathcal{L}_\mathrm{Reg}$ using Eq.~\eqref{eq:final_loss}\;
        }
    }
    Update $\theta$ by minimizing the final objective in Eq.~\eqref{eq:final_loss}\;
}
\end{algorithm}

\subsection{Benchmarks}
\subsubsection{DeCo-Bench.}
\label{app:deco_bench}
\paragraph{Scores.}
DeCo-Bench evaluates multi-image composition primarily from the perspective of \emph{cross-image consistency}, i.e., whether instruction-relevant entities from multiple reference images are correctly and consistently realized in the generated result. It contains settings with $N\in\{2,3,4,5,6\}$ reference images; each setting includes 40 test cases (200 cases in total), and all reference images are real-world photographs.

For each test case, we first crop the instruction-relevant objects from every reference image, and crop the corresponding objects from the generated image. We then score each cropped reference--generation object pair using Gemini-3.1-Pro (the evaluation prompt is provided in Figure~\ref{fig:prompt}). The final score for a sample is computed by averaging all object-level scores:
\begin{equation}
\label{eq:deco_score}
\mathrm{Score}(c)=\frac{1}{|\mathcal{O}_c|}\sum_{o\in\mathcal{O}_c} s(o),
\end{equation}
where $\mathcal{O}_c$ denotes the set of all instruction-relevant objects in case $c$, and $s(o)$ is the Gemini-3.1-Pro score for object $o$ computed from the corresponding cropped reference and generated regions.

\paragraph{Aggregation.}
Let $\mathcal{C}_N$ denote the set of cases with $N$ reference images, and let $\mathcal{C}=\bigcup_{N\in\{2,3,4,5,6\}}\mathcal{C}_N$ denote all cases. We report:
\begin{equation}
\mathrm{Score}_N = \frac{1}{|\mathcal{C}_N|}\sum_{c\in\mathcal{C}_N}\mathrm{Score}(c),\quad
\mathrm{Score}_{\mathrm{overall}} = \frac{1}{|\mathcal{C}|}\sum_{c\in\mathcal{C}}\mathrm{Score}(c).
\end{equation}

\subsubsection{MICo-Bench}
\label{app:mico_bench}
\paragraph{Scores.} MICo-Bench evaluates each multi-image composition case by jointly considering (i) \emph{source preservation} and (ii) \emph{instruction alignment}. Concretely, the final score is computed as:
\begin{equation}
\mathrm{Score} = W \times \mathrm{SC} \times \mathrm{PQ}.
\end{equation}

\paragraph{Tasks.}
MICo-Bench consists of four tasks:
\begin{enumerate}
    \item \textbf{Object-Centric (138 cases):} compose 1--4 distinct objects, or insert 1--2 objects into a semantically compatible scene.
    \item \textbf{Human-Centric (168 cases):} compose multiple persons, or place 1--2 persons into a compatible scene (with explicit gender control).
    \item \textbf{HOI (291 cases):} evaluate human--object interactions, including Person+Object, Person+Clothes, and their combinations.
    \item \textbf{De\&Re (300 cases):} decompose real portraits into human/clothing/object/scene components and then recompose them to assess holistic controllability.
\end{enumerate}

\paragraph{Weight $W$.}
The fraction of source images correctly preserved in the generated output. Faces are verified via ArcFace cosine similarity; non-face elements (objects, clothes, scenes) are verified via a VLM (i.e., Qwen3-VL-30B-A3B-Instruct).

\paragraph{Semantic Consistency (SC).}
To avoid cross-image attention overload, the evaluator compares the generated image against a \emph{human-verified reference image} (instead of jointly attending to all sources) and outputs two sub-scores:
\begin{enumerate}
    \item \textbf{Prompt Following} (PF), measuring compliance with the text instruction (required entities, relations, and scene constraints),
    \item \textbf{Subject Resemblance} (SR), measuring how well the generation matches the reference in subject identity/appearance and semantic content.
\end{enumerate}
We compute:
\begin{equation}
\mathrm{SC}=\sqrt{\mathrm{PF}\times \mathrm{SR}}.
\end{equation}

\paragraph{Perceptual Quality (PQ).}
PQ is evaluated from the generated image alone, including:
\begin{enumerate}
    \item \textbf{Naturalness} (overall realism and coherence of lighting, shadows, geometry, and materials),
    \item \textbf{Artifacts} (absence of visible defects such as distortions, duplicated limbs, and unnatural textures).
\end{enumerate}
We compute:
\begin{equation}
\mathrm{PQ}=\sqrt{\mathrm{Naturalness}\times \mathrm{Artifacts}}.
\end{equation}

\paragraph{Aggregation.}
Let $\mathcal{C}_t$ denote the set of evaluation cases in task $t$, and let $\mathcal{C}=\bigcup_t \mathcal{C}_t$ denote the set of all cases across tasks. We report the task-wise and overall scores as:
\begin{equation}
\mathrm{Score}_t = \frac{1}{|\mathcal{C}_t|}\sum_{c \in \mathcal{C}_t} \mathrm{Score}(c), \quad
\mathrm{Score}_{\mathrm{overall}} = \frac{1}{|\mathcal{C}|}\sum_{c \in \mathcal{C}} \mathrm{Score}(c).
\end{equation}

\subsection{More Visualizations}
\label{sec:app_vis}
\begin{figure}
    \centering
    \includegraphics[width=1\linewidth]{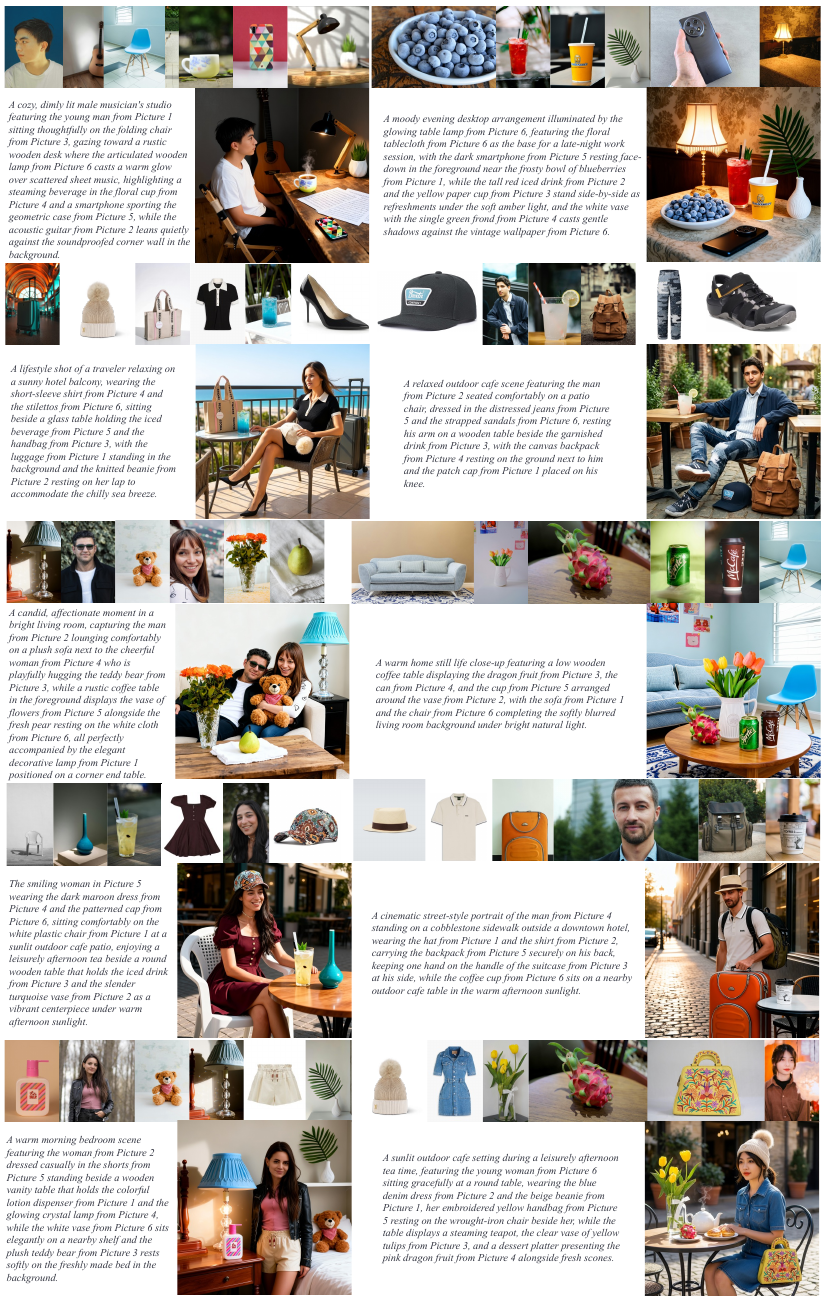}
    \caption{Visualization examples from the DeCo-Bench (6 images).}
    \label{fig:6imgs}
\end{figure}

\begin{figure}
    \centering
    \includegraphics[width=1\linewidth]{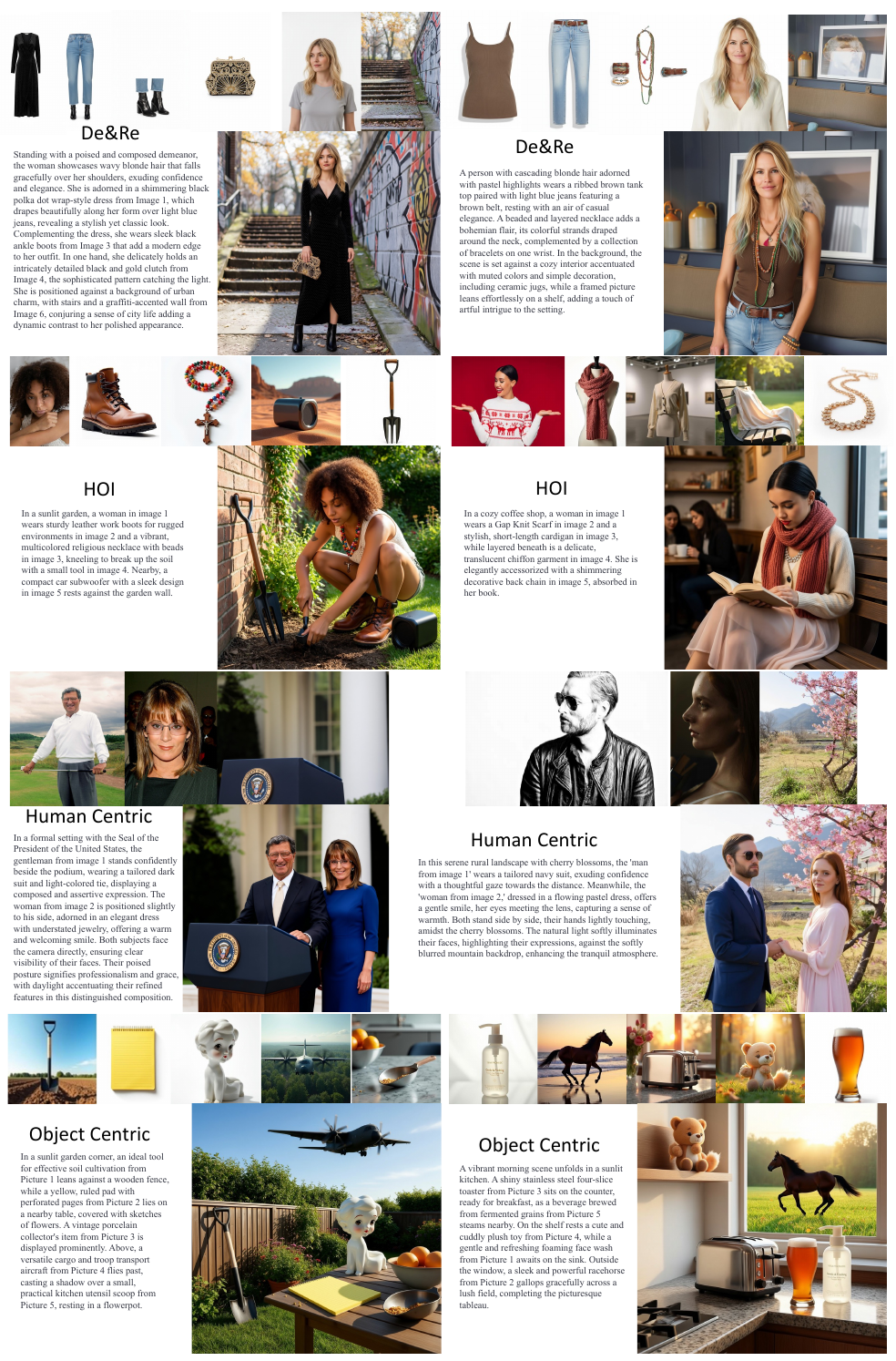}
    \caption{Visualization examples from the MICo-Bench (De\&Re, HOI, Human Centric and Object Centric).}
    \label{fig:mico}
\end{figure}

We provide additional visualization examples on DeCo-Bench (Figure~\ref{fig:6imgs}) and MICo-Bench (Figure~\ref{fig:mico}). As can be seen from these visual results, our method achieves excellent MICo consistency and visual quality.

\begin{figure}
  \centering
  \begin{tcolorbox}[mycodebox, width=\linewidth]
  \begin{lstlisting}
# Role
You are a senior expert in computer vision and image quality assessment, with deep expertise in semantic feature extraction and target consistency analysis.

# Task
Target evaluation label (Label): [obj_label]
Please compare Image A (reference image) and Image B (generated image), and focus your attention exclusively on the entity indicated by the target evaluation label above. Assess the consistency of Image B relative to Image A only for that target.

# Core Constraints
You must strictly follow the principles below for analysis and scoring:

1. Absolute focus (exclusivity): Evaluate strictly and only the target [obj_label] itself. You must completely ignore the background, environment, unrelated people, or any other distracting objects in the scene. Never deduct points because of additions, removals, or substitutions of irrelevant elements.
2. Comparison of major color regions (critically important): No matter what category the target belongs to, you must strictly compare its major color regions.
   - Carefully compare the target's intrinsic dominant colors, color distribution, proportion of color blocks, and key color characteristics.
   - Color changes in core regions, loss of key color blocks, or color inversion must be treated as severe inconsistencies.
3. Generalized feature evaluation dimensions: Based on the semantic category of [obj_label], extract its core features for consistency evaluation:
   - If the target is a living being (such as a person or animal): focus absolutely on biometric identity consistency. Compare only facial feature proportions, facial characteristics, hairstyle/fur texture, face shape/contour, bone structure, and distinctive biological details (such as moles or markings). Ignore clothing, accessories, and body pose changes.
   - If the target is a non-living object (such as an item, product, or building): focus absolutely on physical and design attributes. Compare shape contour, physical structure, material texture, surface patterns, and iconic design elements (such as logos, text, or specific constructions).
4. Exclude physical and viewpoint interference: You must cognitively decouple the target itself from its interaction with the external environment. Compare only the target's intrinsic identity or attributes, and strictly ignore visual differences caused by viewpoint (distance, front/side angle, pitch), pose variation, environmental lighting (brightness, warm/cool light, reflected ambient color), or image noise.

# Output Format
You must output one and only one valid JSON object. Do not include any extra explanation text or Markdown formatting. You must provide the reason first (including analysis of shape/features and major color regions), and then the score (an integer from 0 to 10, where 10 means perfectly consistent and 0 means completely inconsistent). The format must be:

```json
{
  "reason": "Brief explanation of the evaluation basis (must include comparison of core features and major color regions, while excluding viewpoint/lighting interference)",
  "score": "consistency score"
}
  \end{lstlisting}
  \end{tcolorbox}
  \caption{The evaluation prompt (Gemini-3.1-Pro) of DeCo-Bench.}
  \label{fig:prompt}
\end{figure}

\section{Additional Validation and Discussion}
\label{app:additional_validation}

\paragraph{Independent validation of the object-level reward evaluator.}
To assess the trained reward model separately from downstream RL, we evaluated its agreement with human preference labels on EditReward-Bench, which contains approximately 3{,}000 original/edited image pairs. Table~\ref{tab:app_reward_accuracy} reports the accuracy values from this evaluation. The trained evaluator improves upon its untuned counterpart (0.721 versus 0.354), supporting the importance of using a reliable object-level evaluator within the decomposed reward framework. This evaluation measures agreement with the benchmark labels, not reward calibration.
\begin{table}[t]
    \centering
    \caption{Human-label agreement accuracy on EditReward-Bench.}
    \label{tab:app_reward_accuracy}
    \begin{tabular}{lc}
        \toprule
        \textbf{Evaluator} & \textbf{Accuracy$\uparrow$} \\
        \midrule
        GPT-4.1 & 0.602 \\
        GPT-5 & 0.669 \\
        Vanilla reward VLM & 0.354 \\
        Trained reward VLM (ours) & \textbf{0.721} \\
        \bottomrule
    \end{tabular}
\end{table}

\paragraph{Grounding and object matching diagnostics.}
We sampled 10{,}000 examples from the training set, applied Grounding DINO to identify the relevant object regions, and used a VLM to judge whether the detected crops matched the assigned object labels. Under this VLM-based verification protocol, the resulting label--crop matching accuracy was 94.6\%. This result is a diagnostic proxy and should not be interpreted as human-annotated bounding-box localization accuracy or same-category instance-matching accuracy.

\begin{figure}[t]
    \centering
    \includegraphics[width=0.85\linewidth]{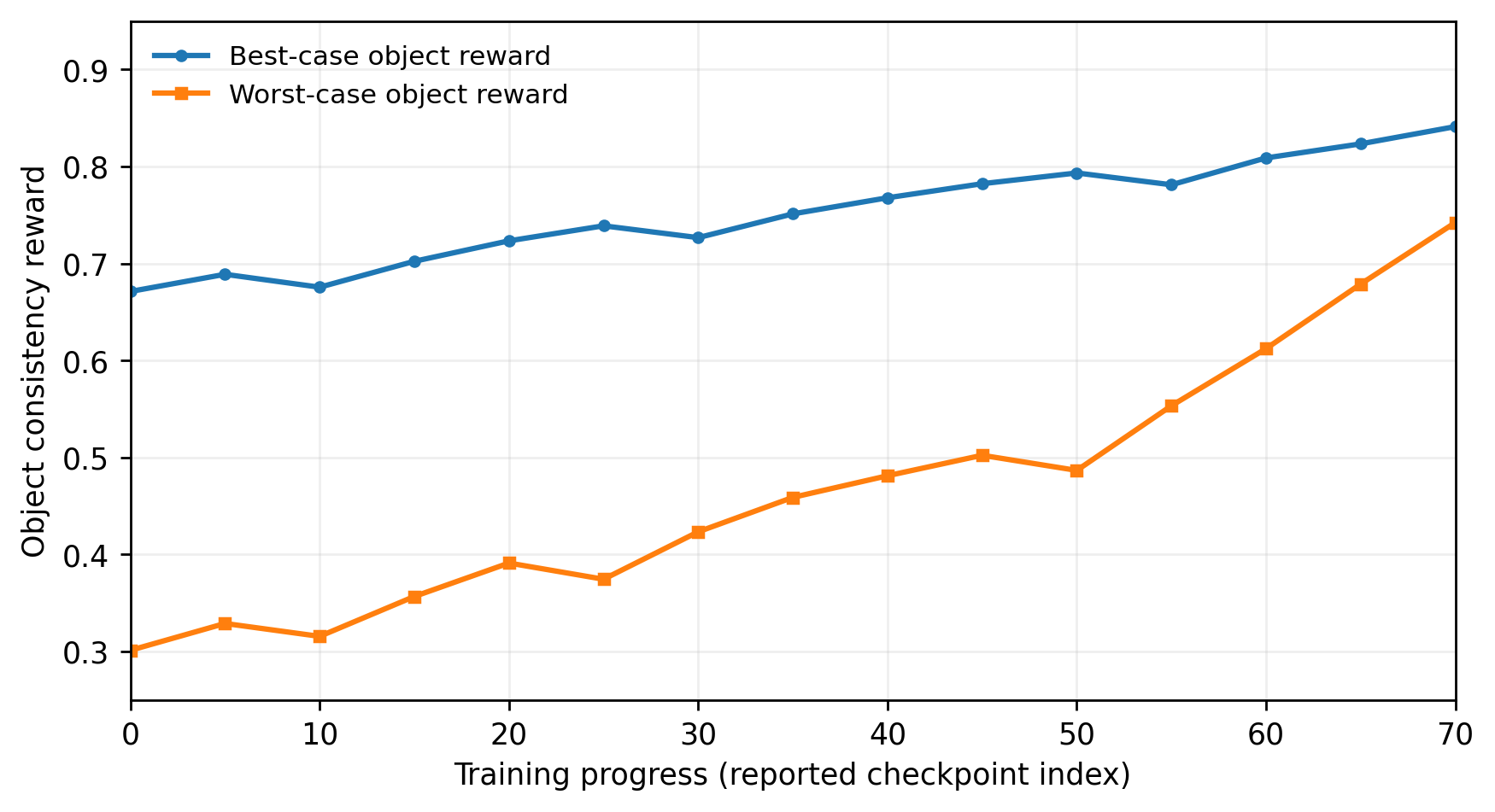}
    \caption{Best-case and worst-case object consistency rewards over the reported training checkpoints.}
    \label{fig:app_object_balance}
\end{figure}

\paragraph{Behavior of best- and worst-case object rewards.}
Figure~\ref{fig:app_object_balance} reports the best-case and worst-case object rewards over the reported training checkpoints. The worst-case reward increases from 0.3012 to 0.7423, while the best-case reward increases from 0.6712 to 0.8412. Thus, both increase under mean aggregation, and the worst-case object shows a larger absolute increase over these checkpoints. These observations are consistent with improvements beyond the easiest objects, although they do not establish that mean aggregation is optimal among possible alternatives.

\paragraph{Evaluation scope and potential reward--metric alignment.}
DeCo-Bench covers 2--6 reference images with 40 test cases per setting (200 cases total). It is designed to assess the localizable object-level consistency targeted by our reward and is therefore an in-domain evaluation. The separately developed MICo-Bench provides complementary evidence beyond this reward-aligned benchmark, including improvements in subject resemblance and perceptual-quality metrics. Nevertheless, this does not replace a dedicated study of possible evaluator bias. The current model was trained for settings with up to six reference images; reliability beyond this range has not been demonstrated.

\paragraph{Choice of RL optimization and auxiliary reward.}
DeCoRL employs Diffusion NFT as a practical post-training optimizer; the object-level reward design is conceptually orthogonal to the choice of diffusion RL algorithm. Early investigations of flow-based GRPO variants suggested higher resource demands and sensitivity to optimization settings, but no matched-training controlled comparison is reported here. The general-purpose auxiliary VLM reward described in Section~\ref{subsec:diffusion_nft} is used to discourage implausible global compositions, including artifacts and incorrect object--subject interactions that a local crop-based score may miss.

\paragraph{Limitations.}
Our decomposed reward primarily captures localizable object identity and appearance. Cropped objects cannot by themselves fully represent abstract styles, spatial relations, or interaction and binding constraints. Grounding and same-category object matching can fail, while VLM-based evaluation may introduce bias or be vulnerable to reward hacking. DeCo-Bench is relatively small, and the main experiments use one editing backbone and 2--6 reference images. The pipeline also relies on multiple large models, increasing computational and reproducibility costs. A controlled holistic-versus-object-level reward comparison under matched RL training remains an important avenue for future work.


\end{document}